\pdfoutput=1
\documentclass[11pt]{article}

\usepackage[]{ACL2023}

\usepackage{times}
\usepackage{latexsym}
\usepackage{diagbox}

\usepackage[utf8]{inputenc}
\usepackage{graphicx}
\usepackage{pgfplots}
\usepackage{booktabs}
\usepackage{float}
\usepackage{listings}
\usepackage{tabularx}
\usepackage{booktabs} % For better table lines
\usepackage{array} % For table alignment

\usepackage[T1]{fontenc}
\usepackage[utf8]{inputenc}

\usepackage{microtype}

\usepackage{url}            % simple URL 
\usepackage{microtype}
\usepackage{amsmath}
\usepackage{amssymb}
\usepackage {tcolorbox}
\tcbuselibrary{theorems}
\usepackage{graphicx}
\usepackage{comment}
\usepackage{array}
\usepackage{mathtools}
\usepackage{csquotes}
\usepackage{multirow}
\usepackage{enumitem}

\title{ReadCtrl: Personalizing text generation with readability-controlled instruction learning}

\author{Hieu Tran \thanks{* indicates equal contribution} $^1$, Zonghai Yao \footnotemark[1] $^1$, Lingxi Li  $^1$, Hong Yu$^{1, 2, 3, 4}$\\
$^1$ Manning College of Information and Computer Sciences, University of Massachusetts Amherst, MA, USA\\
    $^2$ Department of Medicine, University of Massachusetts Medical School, Worcester, MA, USA\\
    $^3$ Miner School of Computer and Information Sciences, University of Massachusetts Lowell, MA, USA\\
    $^4$ Center for Healthcare Organization and Implementation Research, VA Bedford Health Care, MA, USA\\
  {\tt \{hieutran, zonghaiyao\}@umass.edu}\\ 
  \\
}

\begin{document}
\maketitle
\begin{abstract}
Content generation conditioning on users' readability is an important application for personalization.
In an era of large language models (LLMs), readability-controlled text generation based on LLMs has become increasingly important. 
This paper introduces a novel methodology called ``Readability-Controlled Instruction Learning (ReadCtrl),'' 
which aims to instruction-tune LLMs to tailor users' readability levels.
Unlike the traditional methods, which primarily focused on categorical readability adjustments—typically classified as high, medium, and low or expert and layperson levels—with limited success, ReadCtrl introduces a dynamic framework that enables LLMs to generate content at various (near continous level) complexity levels, thereby enhancing their versatility across different applications.
Our results show that the ReadCtrl-Mistral-7b models significantly outperformed strong baseline models such as GPT-4 and Claude-3, with a win rate of 52.1\%:35.7\% against GPT-4 in human evaluations.
Furthermore, ReadCtrl has shown significant improvements in automatic evaluations, as evidenced by better readability metrics (e.g., FOG, FKGL) and generation quality metrics (e.g., BLEU, SARI, SummaC-Factuality,  UniEval-Consistency and Coherence).
% Furthermore, ReadCtrl has demonstrated substantial enhancements across multiple dimensions in automatic evaluations, incorporating both readability metrics (e.g., FOG and FKGL) and generation quality metrics (e.g., BLEU, SARI, Factuality, Consistency, Coherence, and LLM-as-judge evaluation).
These results underscore ReadCtrl's effectiveness and tenacity in producing high-quality, contextually appropriate outputs that closely align with targeted readability levels, marking a significant advancement in personalized content generation using LLMs.
% \textcolor{red}{HONG: I would like you to output some results as you have an excellent result}
% ~\footnote{We will release prompts, codes, and dataset upon acceptance.}.
~\footnote{Our code and data will be released at \url{https://github.com/bio-nlp/ReadCtrl}}.
\end{abstract}

\section{Introduction}
\label{Intro}

\begin{figure}[!h]
    \vspace{-2mm}
    \centering
    \includegraphics[width=1\linewidth]{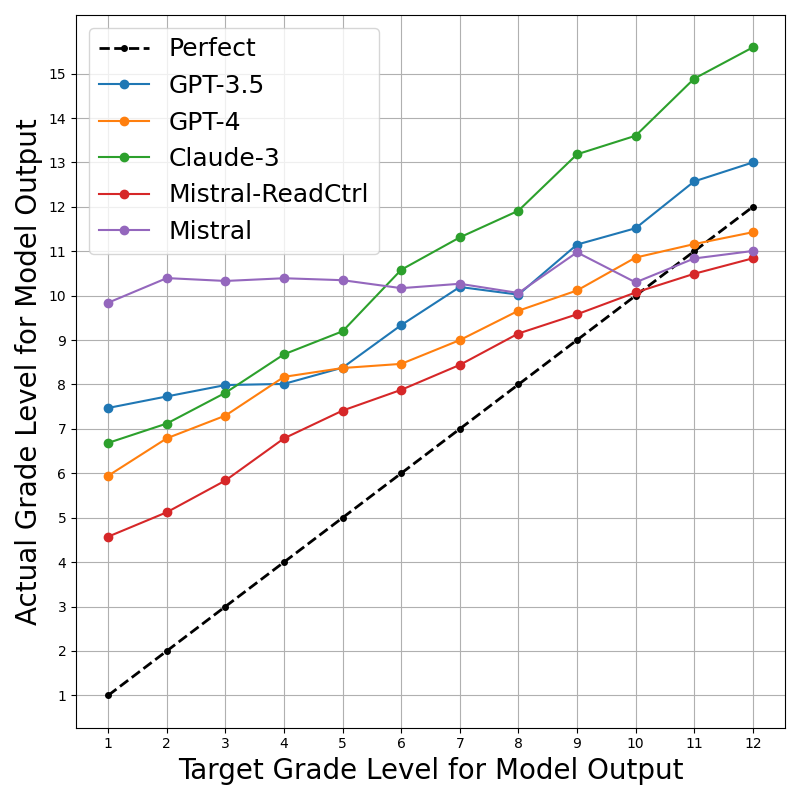}
    \vspace{-6mm}
    \caption{ReadCtrl instruction following ability. 
    While current SOTA LLMs such as GPT and Claude (under the few-shot setting) show an upward trend in aligning their output with the target grade level, they fall significantly short of the `perfect' adherence curve. 
    Other weaker LLMs like Mistral-7b demonstrate little to no capacity to adjust to ReadCtrl instructions, as indicated by the flat line parallel to the x-axis. 
    Notably, Mistral-ReadCtrl's performance closely approaches `perfect', showcasing its advanced capability to tailor output to the specified readability level as set out by ReadCtrl instructions.}
    \label{fig:ReadCtrl_instruction_following_ability}
    \vspace{-6mm}
\end{figure}

Existing personalization methods mainly focus on the semantics of the content that individuals need, such as retrieving information based on individuals' search queries~\cite{Chen2023WhenLL,kirk2024benefits,shanahan2023role} and summarization based on content representation~\cite{richardson2023integrating}.
However, one important aspect of personalization that has not been widely explored is readability-controlled content generation~\cite{vajjala2021trends}. 
This involves tailoring content to match individuals' readability levels, which can vary widely due to differences in education and related experiences and training~\cite{ribeiro2023generating}.
The emergence of large language models (LLMs) has further advanced this field, ushering in a transformative era of automatic content generation~\cite{pu2023chatgpt}.
It is crucial for content generated by these models to be accurate, relevant, and consistent with the cognitive abilities of the target audience. 
The emphasis on customized content creation underscores the critical role of customization and personalization in digital interactions, especially in environments where explicit instructions (typically from LLMs' users) must be strictly followed~\cite{zhou2023instruction,sun2023evaluating,qin2024infobench}.
At the heart of this innovative area are the principles of readability control instructions, designed to dynamically adapt the output vocabulary distribution to the specific context of each interaction. 
This can be achieved by analyzing chat history, interpreting user profiles, or responding to direct interaction requests, significantly enhancing LLMs' versatility~\cite{huang2023recommender}.

Previous efforts in the domain of controllable text generation have primarily centered on binary readability adjustments, typically categorized into tasks of simplification, complication, or sibling style transfer~\cite{goldsack2022making,guo2021automated,luo2022readability}.
Despite their objectives, these approaches often fail to fully address the diverse personalization needs due to the limited variety in training data and a concentrated emphasis on readability.
In traditional supervised fine-tuning scenarios, designing multiple readability-level ground truths for training data to facilitate readability control is not scalable.
As a result, models may not have sufficient exposure to varied text complexities, limiting their ability to adjust content according to user-specific readability needs accurately.
In response, the field of artificial intelligence is shifting towards more dynamic systems that can adapt outputs to meet users' unique preferences and requirements~\cite{kirk2024benefits}.
This shift is heralding a new era of personalized content creation that extends beyond standard domains like information retrieval to specialized areas, enhancing user engagement and satisfaction through tailored content.

This paper addresses these challenges by introducing a novel methodology termed ``readability-controlled instruction learning (ReadCtrl).''
Our findings demonstrate that ReadCtrl can empower LLMs to transform input text into content with controlled readability accurately. 
As illustrated in Figure~\ref{fig:ReadCtrl_instruction_following_ability}, our investigation across a range of state-of-the-art LLMs shows varying degrees of compliance with readability-controlled instructions. 
Mainstream models like GPT~\cite{achiam2023gpt} and Claude~\cite{TheC3}, despite demonstrating an Upward trend, fall significantly short of the ideal `perfect' adherence curve—they show a tendency towards compliance but lack precise output control.
In contrast, models that previously struggled with readability control, such as Mistral-7b~\cite{jiang2023mistral}—illustrated almost as a horizontal line in the figure—have shown significant enhancement with the integration of ReadCtrl, such as Mistral-ReadCtrl. 
These models now nearly meet the ideal performance curve, showcasing their improved ability to customize outputs to specific readability demands.
Specifically, ReadCtrl incorporates explicit instruction tuning based on readability and is rigorously tested through tasks designed to evaluate the model’s ability to adjust output complexity. 
These tasks include text simplification, which aims to reduce the output's readability relative to the input; paraphrase generation, which maintains the input's readability; and semantic entailment generation, which adjusts the output’s readability, potentially increasing or decreasing it in relation to the input. 
By employing a clear instruction—``\emph{Given an input text, please output an entailment with a readability score around \{target readability score\}}''—models like Mistral-ReadCtrl demonstrate the efficacy of ReadCtrl in generating content that is not only semantically coherent but also finely adjusted to meet diverse comprehension needs and preferences.

In our initial experiments, we evaluated the model's performance in a "seen setting," where models were tested using datasets on which they were trained, such as ASSET~\cite{alva2020asset} for text simplification, PAWS~\cite{zhang2019paws} for paraphrase generation, and SNLI~\cite{bowman2015large} for semantic entailment. This setting established a baseline for instruction tuning, confirming that the models could effectively adhere to readability instructions in familiar contexts. Subsequent experiments involved an "unseen setting," testing the models against new datasets they had not encountered during training, such as WikiSmall~\cite{zhu2010monolingual} for text simplification, MRPC~\cite{dolan2005automatically} for paraphrase generation, and MultiNLI~\cite{williams2017broad} for semantic entailment. This phase was critical for assessing the models' adaptability and accuracy in novel contexts, reflecting their generalizability and real-world applicability.
We utilized several readability assessment metrics, including the Gunning Fog Index~\cite{gunning1952technique} and Flesch-Kincaid Grade Level~\cite{kincaid1975derivation}, to quantitatively evaluate the complexity of the generated texts. Additionally, we employed a range of automatic generation metrics for generation quality evaluation, such as BLEU~\cite{papineni2002bleu}, SARI~\cite{xu2016optimizing}, Factuality~\cite{laban2022summac}, Consistency and Coherence~\cite{zhong2022towards}, to assess the quality of the generated texts, aiming to balance readability, information retention, factuality, consistency, and coherence during evaluation.

These evaluations confirmed the effectiveness of our ReadCtrl methodology across a diverse range of tasks and datasets. Particularly, Mistral-ReadCtrl excelled in both seen and unseen settings, showcasing robust performance metrics. For instance, in the unseen MRPC dataset, Mistral ReadCtrl achieved the lowest readability gap (1.66), the highest factuality (0.8184), and excellent BLEU (0.3798) and SARI (44.4327) scores, significantly outperforming competitors like GPT-4 and Claude-3. In the WikiSmall dataset, it recorded a readability gap of just 2.09, the highest coherence score (0.9763), and a strong SARI score of 42.1033.
Furthermore, detailed human and LLM-as-a-judge~\cite{lan2024criticbench} evaluations were conducted to compare Mistral-ReadCtrl with GPT-4 across different tasks and readability requirements. 
Mistral-ReadCtrl demonstrated superior performance, achieving a win rate of 52.1\% in human evaluations and 58.3\% in AI assessments, compared to GPT-4's 35.7\% and 38.4\%, respectively. 
Notably strong results were observed in tasks involving WikiSmall (62.5\% in Human Eval, 67.7\% in AI Eval) and SNLI (66.7\% in Human Eval).

% These results underscore Mistral-ReadCtrl’s ability to produce high-quality, contextually appropriate outputs that closely align with targeted readability levels, highlighting its versatility and robustness for various NLP applications.

\begin{figure*}
    \centering
    \includegraphics[width=1.0\textwidth]{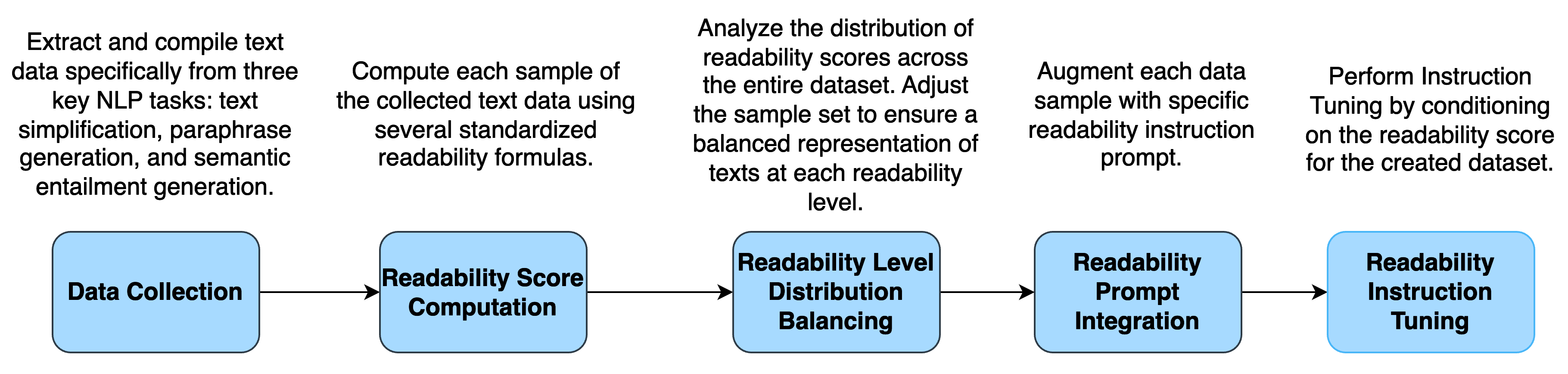}
    \caption{Overview of ReadCtrl data construction.}
    \vspace{-4mm}
    \label{fig:method_overview}
\end{figure*}

% \section{Problem Statement}
% \label{Sec:problem_statement}

\section{Methodology}
\subsection{Task Overview}

Our methodology is designed to evaluate the effectiveness of instruction tuning conditional on readability across a suite of tasks, specifically focusing on text simplifications, paraphrase generation, and semantic entailment generation. These tasks are strategically chosen to test the model's capability in adjusting the complexity of its output to match specified readability levels. They serve a broad spectrum of applications, from enhancing educational material accessibility to refining technical documentation for diverse audiences.

\begin{itemize}
[leftmargin=.1in,topsep=0.3pt]
\setlength\itemsep{0em}
\vspace{-0.2em}
    \item \textbf{Text Simplifications:} Here, the aim is to reduce the readability level of the given input text, making it more accessible to a wider audience or readers with varying comprehension skills. This task challenges the model to simplify complex text while preserving its essential content and meaning, demonstrating the ability to decrease textual complexity upon demand.
        
    \item \textbf{Paraphrase Generation:} In this task, the model is tasked with rewording the given text to produce a paraphrase that maintains the original's readability level. This requires a nuanced understanding of language to ensure the output remains true to the input's complexity and style, facilitating content reformulation without altering its accessibility.
    
    \item \textbf{Semantic Entailment Generation:} This involves creating text that semantically follows from the given input, with the flexibility to increase or decrease the readability level. The model must grasp the underlying meaning of the input text and generate output that logically entails the input, demonstrating versatility in producing content with adjustable complexity levels.

\end{itemize}

We employ the instruction tuning approach conditional on readability for all these tasks. This method provides explicit instructions to the model to control the output text's readability score, ensuring that the generated content aligns with the intended complexity level for the target audience. This approach underlines our belief that these tasks can all contribute to readability control generation, where, depending on the task—be it text simplification, paraphrase generation, or semantic entailment generation—the model is calibrated to generate output with the desired readability level. In text simplification, the goal is to lower the readability of the output relative to the input, while in paraphrase generation, the output's readability should mirror the input's. For the semantic entailment generation task, the output's readability may vary, being either higher or lower than the input's, thereby offering a versatile tool for adjusting text complexity across a wide range of contexts.

\subsection{Instruction Design for Readability Control}

To achieve the desired readability level across various tasks, we employ straightforward and singular instruction. This approach emphasizes the model's ability to tailor its output to meet specific readability goals, demonstrating its versatility and effectiveness in readability control. The instruction is as follows:

"Given an input text, please output an entailment with a readability score around {target readability score}."

This concise instruction mandates the model to generate content that not only semantically follows from the given input but also aligns with a specified readability level, showcasing the model's capacity to produce targeted outputs that cater to diverse comprehension needs and preferences.
% The bidirectional approach enables the flexibility of adjusting text complexity based on the target audience's needs, making it suitable for a broad range of applications beyond medical text simplification.

\subsection{Implementation and Readability Scoring}

The readability of the generated text is quantitatively evaluated using a suite of established readability metrics. We calculate the following readability scores~\footnote{More details can be found in Appendix~\ref{apx:read_metrics}.}:

\begin{itemize}
[leftmargin=.1in,topsep=0.3pt]
\setlength\itemsep{0em}
\vspace{-0.2em}
    \item \textbf{Gunning Fog Index:} Estimates the years of formal education required to understand the text on the first reading~\cite{gunning1952technique}.
    \item \textbf{Flesch-Kincaid Grade Level:} Translates the US grade level needed to comprehend the text~\cite{kincaid1975derivation}.
    \item \textbf{Automated Readability Index:} Outputs a score correlating to the US grade level necessary for understanding~\cite{senter1967automated}.
    \item \textbf{Coleman-Liau Index:} Estimates the US grade level needed to comprehend the text using letter count instead of syllable count~\cite{coleman1975computer}.
\end{itemize}

These metrics are selected for their diverse approaches to assessing text complexity, offering a comprehensive understanding of the text's readability. Subsequently, an average Reading Grade Level (RGL) is derived from these scores to represent the text's overall readability. The integration of these readability assessments into our methodology allows a nuanced approach to generating text that meets the specified readability criteria. By adjusting the instruction based on the target RGL, we can fine-tune the complexity of the output, making our approach adaptable to a wide range of applications, from educational content to technical documentation. This process underscores the importance of readability in tailoring content to specific audience needs, a critical factor in communication effectiveness across various domains.

\section{Experiments}

\subsection{Dataset}
Our experimental framework is designed to assess the model's performance across various tasks, specifically focusing on text simplification, paraphrase generation, and semantic entailment generation. To facilitate a comprehensive evaluation, we utilize six distinct datasets, two for each task, which enables us to explore the model's capabilities in both seen and unseen settings. The datasets employed in our experiments are outlined as follows:

\begin{itemize}
[leftmargin=.1in,topsep=0.5pt]
\setlength\itemsep{0em}
\vspace{-0.2em}
    \item \textbf{Text Simplification:} For this task, we use the ASSET \cite{alva2020asset} and WikiSmall \cite{zhu2010monolingual} datasets. ASSET is a diverse corpus for automatic sentence simplification, providing high-quality simplifications with multiple references per source sentence, making it ideal for instruction tuning and evaluation in seen settings. WikiSmall serves as an additional dataset for evaluating performance in an unseen setting, offering a different collection of simplified sentences derived from Wikipedia articles.
    
    \item \textbf{Paraphrase Generation:} We utilize the PAWS \cite{zhang2019paws} (Paraphrase Adversaries from Word Scrambling) and MRPC (Microsoft Research Paraphrase Corpus) \cite{dolan2005automatically} datasets. PAWS contains pairs of sentences paraphrasing each other, including those constructed through controlled word scrambling, making it suitable for training and the seen setting evaluations. MRPC offers a collection of sentence pairs labeled as paraphrases or not, sourced from online news sources, to test the model's paraphrasing ability in unseen settings.
    
    \item \textbf{Semantic Entailment Generation:} For this task, the SNLI (Stanford Natural Language Inference) \cite{bowman2015large} and MultiNLI (Multi-Genre Natural Language Inference) \cite{williams2017broad} datasets are employed. SNLI is a large collection of sentence pairs annotated with textual entailment information, used for instruction tuning and seen setting evaluation. MultiNLI extends this to a broader range of genres and contexts, providing a robust challenge for the model in unseen settings.

\end{itemize}

In our experimental setup, instruction tuning is performed on the training sets of ASSET, PAWS, and SNLI to align the model's output with specific readability goals. The effectiveness of this approach is then evaluated in two distinct settings: a \textit{seen setting}, using the test sets of ASSET, PAWS, and SNLI, and an \textit{unseen setting}, using the test sets of WikiSmall, MRPC, and MultiNLI. This methodology allows us to not only measure the model's immediate response to the instruction tuning but also its generalizability and adaptability to different textual contexts and tasks.

\subsection{Evaluation Metrics}

To comprehensively evaluate the model's performance across the different tasks, we employ a multifaceted set of metrics that assess various aspects of the generated texts. These metrics enable us to gauge the model's effectiveness in adjusting readability, maintaining factual accuracy, and ensuring textual coherence and consistency. The following metrics are used:

\begin{itemize}
[leftmargin=.1in,topsep=0.3pt]
\setlength\itemsep{0em}
\vspace{-0.2em}
    \item \textbf{Average Readability Score:} This metric calculates the average readability level of the generated texts, providing insight into the overall accessibility of the content produced by the model.
    
    \item \textbf{Readability Gap (Delta):} The readability gap is measured as the difference between the requested readability level and the actual readability level of the generated text. This metric assesses the model's precision in hitting target readability levels.
    
    \item \textbf{Factuality:} Factuality is evaluated based on the methodology from the SummaC \cite{laban2022summac} work, which offers a means to assess the factual alignment of the generated text with the source content or input.
    
    \item \textbf{Consistency and Coherence:} These aspects are measured using criteria from the UniEval \cite{zhong2022towards} framework, which provides standardized metrics for evaluating the logical consistency and coherence of the text, ensuring that the generated content is not only readable but also logically structured and coherent.
    
    \item \textbf{SARI:} The SARI (System output Against References and the Input sentence) \cite{xu2016optimizing} metric is utilized to assess the quality of text simplification. It measures the model's ability to produce simplified text that is both accurate and helpful, comparing the generated output against both the original text and reference simplifications.
    
    \item \textbf{BLEU:} The BLEU (Bilingual Evaluation Understudy) \cite{papineni2002bleu} metric is applied to evaluate paraphrase generation and semantic entailment tasks. It quantifies the linguistic similarity between the generated texts and reference texts, indicating the model's capability to produce coherent and contextually appropriate content.
\end{itemize}

These metrics collectively offer a robust framework for assessing the nuanced performance of the model across various dimensions of text generation, readability adjustment, and content quality.

\begin{table*}[h!]
    \centering
    \scalebox{0.78}{
    \begin{tabular}{lcccccc}
        \hline
Models & Readability Gap$\downarrow$ & Factuality$\uparrow$ & Consistency$\uparrow$ &	Coherence$\uparrow$	& BLEU$\uparrow$ & SARI$\uparrow$  \\
\hline
\multicolumn{7}{c}{\textbf{ASSET (seen) | WikiSmall (unseen) - Text Simplification}} \\
Claude-3 & 3.6323 | 4.53 & 0.5221 |  0.4612 & 0.9301 | 0.9391	& 0.934 | 0.9396 & 0.1874 | 0.1606 & 40.6964 | 32.9996 \\

GPT-3.5 & 2.8635 | 3.12 & 0.7231 | 0.6721 & 0.9641 | 0.9401 & 0.9648 | 0.9231 & 0.2739 | 0.194 & 41.0061 | 33.9842 \\

GPT-4 &  2.7465 | 2.69 & 0.6547 | 0.5892 & \textcolor{red}{\textbf{0.9688}} | \textcolor{red}{\textbf{0.9556}} & \textcolor{red}{\textbf{0.9687}} | 0.949 & 0.2061 | 0.1666 & 39.7319 | 31.4657 \\

Mistral-ReadCtrl & \textcolor{red}{\textbf{1.8384}} | \textcolor{red}{\textbf{2.09}} & \textcolor{red}{\textbf{0.7687}} | \textcolor{red}{\textbf{0.7168}} & 0.9423 | 0.9477 & 0.9653 | \textcolor{red}{\textbf{0.9763}} & \textcolor{red}{\textbf{0.4317}} | \textcolor{red}{\textbf{0.4321}} & \textcolor{red}{\textbf{49.3521}} | \textcolor{red}{\textbf{42.1033}} \\
\hline

\multicolumn{7}{c}{\textbf{SNLI (seen) | MultiNLI (unseen) - Semantic Entailment Generation}} \\
Claude-3 &  4.6433 | 5.64 & 0.5102 | 0.3904 & 0.919 | 0.8292	& 0.9331 | 0.8346 & 0.0446 | 0.0303 & 48.3281 | 44.4344 \\

GPT-3.5 & 2.8333 | 6.7 & 0.5176 | 0.3967 & 0.9049 | 0.8829 & 0.8982 | 0.896	& 0.0875 | 0.0378 & 51.0201 | 44.0607 \\

GPT-4 &  2.4733 | 3.36	& 0.5632 | 0.5167 &	0.9488 | \textcolor{red}{\textbf{0.8961}} & 0.9382 | 0.8879 &	0.105 | 0.0562 & \textcolor{red}{\textbf{52.1153}} | \textcolor{red}{\textbf{46.4204}} \\

Mistral-ReadCtrl &  \textcolor{red}{\textbf{1.8733}} | \textcolor{red}{\textbf{2.21}} & \textcolor{red}{\textbf{0.7406}} | \textcolor{red}{\textbf{0.6542}} & \textcolor{red}{\textbf{0.9491}} | 0.8804 & \textcolor{red}{\textbf{0.9437}} | \textcolor{red}{\textbf{0.9122}} & \textcolor{red}{\textbf{0.183}} | \textcolor{red}{\textbf{0.1137}} & 51.6644 | 43.8289 \\
\hline

\multicolumn{7}{c}{\textbf{PAWS (seen) | MRPC (unseen) - Paraphrase Generation}} \\
Claude-3 &  2.4333 | 2.61 & 0.5141 | 0.4736 & 0.921 | 0.9154 & 0.9183 | 0.9012 & 0.2393 | 0.1679 & 38.3459 | 36.7783 \\

GPT-3.5 & 1.5433 | 2.64 & 0.7443 | 0.5868 & 0.9761 | \textcolor{red}{\textbf{0.9683}} & 0.9746 | 0.9679 & 0.3873 | 0.2059 & 37.9808 | 37.3417 \\

GPT-4 & 1.4467 | 2.19 & 0.7085 | 0.5203 & \textcolor{red}{\textbf{0.979}} | 0.9635 & \textcolor{red}{\textbf{0.978}} | 0.9639 & 0.3122 | 0.153 & 34.3525 | 34.8477 \\

Mistral-ReadCtrl & \textcolor{red}{\textbf{0.6367}} | \textcolor{red}{\textbf{1.66}} & \textcolor{red}{\textbf{0.7871}} | \textcolor{red}{\textbf{0.8184}} & 0.9677 | 0.9669 & 0.9735 | \textcolor{red}{\textbf{0.9769}} & \textcolor{red}{\textbf{0.6649}} | \textcolor{red}{\textbf{0.3798}} & \textcolor{red}{\textbf{60.5332}} | \textcolor{red}{\textbf{44.4327}} \\
\hline
\end{tabular}}
\vspace{-2mm}
\caption{Main results for seen | unseen tasks in ReadCtrl.}
\vspace{-4mm}
\label{tab:seen_unseen}
\end{table*}

% \begin{table*}[h!]
%     \centering
%     \scalebox{0.9}{
%     \begin{tabular}{lcccccc}
%         \hline
% Models & Readability Gap$\downarrow$ & Factuality$\uparrow$ & Consistency$\uparrow$ &	Coherence$\uparrow$	& BLEU$\uparrow$ & SARI$\uparrow$  \\
% \hline
% \multicolumn{7}{c}{ASSET - Text Simplification} \\
% Claude-3 & 3.6323	& 0.5221 & 0.9301	& 0.934	& 0.1874 & 40.6964 \\
% GPT-3.5 & 2.8635 & 0.7231 & 0.9641 & 0.9648	& 0.2739 & 41.0061 \\
% GPT-4 &  2.7465 & 0.6547 & \textbf{0.9688} &	\textbf{0.9687} & 0.2061 & 39.7319 \\
% Mistral-ReadCtrl &  \textbf{1.8384} &	\textbf{0.7687} & 0.9423	& 0.9653 &	\textbf{0.4317} &	\textbf{49.3521} \\
% \hline

% \multicolumn{7}{c}{SNLI - Semantic Entailment Generation} \\
% Claude-3 &  4.6433 & 0.5102	& 0.919	& 0.9331	& 0.0446 & 48.3281 \\
% GPT-3.5 & 2.8333 &	0.5176	& 0.9049 &	0.8982	& 0.0875	& 51.0201 \\
% GPT-4 &  2.4733	& 0.5632 &	0.9488 &	0.9382 &	0.105	& \textbf{52.1153} \\
% Mistral-ReadCtrl &  \textbf{1.8733} &	\textbf{0.7406}	& \textbf{0.9491} &	\textbf{0.9437}	& \textbf{0.183} & 51.6644 \\
% \hline

% \multicolumn{7}{c}{PAWS - Paraphrase Generation} \\
% Claude-3 &  2.4333 & 0.5141 & 0.921 & 0.9183 & 0.2393 &	38.3459 \\
% GPT-3.5 & 1.5433 & 0.7443 &	0.9761 & 0.9746 & 0.3873 & 37.9808 \\
% GPT-4 & 1.4467 & 0.7085 & \textbf{0.979} & \textbf{0.978} & 0.3122 & 34.3525 \\
% Mistral-ReadCtrl & \textbf{0.6367} &	\textbf{0.7871} & 0.9677 & 0.9735 & \textbf{0.6649} &	\textbf{60.5332} \\
% \hline
% \end{tabular}}
% \caption{Performance on seen tasks}
% \label{tab:seen}
%     % \vspace{-4mm}
% \end{table*}

\subsection{Evaluated Models}
In our study, we evaluate a diverse set of models to understand their efficacy in handling tasks related to term definition generation, text simplification, and text complication, particularly focusing on adjusting text complexity according to specified readability levels. The models include:

\begin{itemize}
[leftmargin=.1in,topsep=0.3pt]
\setlength\itemsep{0em}
\vspace{-0.2em}
    \item \textbf{GPT-3.5:} As a precursor to GPT-4, GPT-3.5 has demonstrated substantial capabilities in generating human-like text across various tasks. It serves as a baseline to understand the incremental improvements brought about by its successors and other models.
    
    \item \textbf{GPT-4:} The latest iteration from OpenAI's GPT series at the time of our study, GPT-4, represents a significant leap in language model performance, offering improved comprehension and generation capabilities over its predecessors.
    
    \item \textbf{Claude-3:} As a model known for its understanding and generation abilities, Claude-3 has been included as a baseline for its efficiency in handling various NLP tasks and its purported adaptability to instruction-based prompts, making it a relevant comparison for our instruction-tuned model.
    
    \item \textbf{Mistral 7B ReadCtrl:} Our proposed model has been instruction-tuned to adjust the readability level of generated texts based on explicit instructions. Mistral 7B is designed to excel in the specific tasks of text simplification, paraphrase generation, and semantic entailment generation, leveraging instruction tuning to achieve precise control over the readability of its outputs.
\end{itemize}

Each of these models brings unique strengths and capabilities to the table, allowing us to conduct a comprehensive comparison that not only highlights Mistral 7B's advancements in controlling readability but also situates these achievements within the broader context of current NLP technologies. By evaluating Mistral 7B against these established models, we aim to demonstrate its efficacy and potential applications in enhancing readability control in automatic text generation.\textbf{}

% % \begin{table*}[h!]
% %     \centering
% %     \scalebox{0.9}{
% %     \begin{tabular}{lcccccc}
% %         \hline
% % Models & Readability Gap$\downarrow$ & Factuality$\uparrow$ & Consistency$\uparrow$ &	Coherence$\uparrow$	& BLEU$\uparrow$ & SARI$\uparrow$  \\
% % \hline
% % \multicolumn{7}{c}{WikiSmall - Text Simplification} \\
% % Claude-3 & 4.53	& 0.4612 &	0.9391 & 0.9396 & 0.1606 & 32.9996 \\
% % GPT-3.5 & 3.12 & 0.6721	& 0.9401 & 0.9231 &	0.194 &	33.9842 \\
% % GPT-4 &  2.69 & 0.5892 & \textbf{0.9556} & 0.949 & 0.1666 & 31.4657 \\
% % Mistral-ReadCtrl & \textbf{2.09} & \textbf{0.7168} & 0.9477 & \textbf{0.9763} &	\textbf{0.4321} & \textbf{42.1033} \\
% % \hline

% % \multicolumn{7}{c}{MultiNLI - Semantic Entailment Generation} \\
% % Claude-3 & 5.64 & 0.3904 & 0.8292 &	0.8346 & 0.0303 & 44.4344 \\
% % GPT-3.5 & 6.7 &	0.3967 & 0.8829 & 0.896 & 0.0378 & 44.0607 \\
% % GPT-4 & 3.36 & 0.5167 &	\textbf{0.8961} & 0.8879 & 0.0562 & \textbf{46.4204} \\
% % Mistral-ReadCtrl & \textbf{2.21} & \textbf{0.6542} & 0.8804 & \textbf{0.9122} &	\textbf{0.1137} & 43.8289 \\
% % \hline

% % \multicolumn{7}{c}{MRPC - Paraphrase Generation} \\
% % Claude-3 &  2.6067 & 0.4736 & 0.9154 & 0.9012 &	0.1679 & 36.7783 \\
% % GPT-3.5 & 2.6367 & 0.5868 & \textbf{0.9683} & 0.9679	& 0.2059 & 37.3417 \\
% % GPT-4 & 2.1867 & 0.5203 & 0.9635 & 0.9639 &	0.153 & 34.8477 \\
% % Mistral-ReadCtrl & \textbf{1.66} & \textbf{0.8184}	& 0.9669 & \textbf{0.9769} &	\textbf{0.3798} & \textbf{44.4327} \\
% % \hline

% \end{tabular}}
% \caption{Performance on unseen tasks}
% \label{tab:unseen}
%     % \vspace{-4mm}
% \end{table*}

\subsection{Results}

\subsubsection{Performance on seen tasks}
Table \ref{tab:seen_unseen} presents a performance comparison of Claude-3, GPT-3.5, GPT-4, and our model, Mistral-ReadCtrl, on seen tasks involving three datasets where instruction tuning was implemented: ASSET, SNLI, and PAWS. Regarding the Readability Gap, Mistral-ReadCtrl demonstrates superior precision in adhering to target readability levels, achieving the lowest scores across all datasets, indicating effective control over text readability. Factuality scores, which assess the accuracy of content compared to the original, show that Mistral-ReadCtrl maintains higher factual consistency than its counterparts. When evaluating Consistency and Coherence, which measure the logical flow and structural soundness of texts, Mistral-ReadCtrl performs robustly, though it is slightly outperformed by GPT-4 in the PAWS dataset. For BLEU and SARI metrics, critical for evaluating the linguistic and contextual appropriateness of text simplification and paraphrase generation, Mistral-ReadCtrl again posts the highest scores, showcasing its efficacy in producing high-quality, contextually appropriate responses.

\subsubsection{Performance on unseen tasks}
Table \ref{tab:seen_unseen} illustrates the performance of four models — Claude-3, GPT-3.5, GPT-4, and Mistral-ReadCtrl — on unseen tasks, using the datasets WikiSmall for text simplification, MultiNLI for semantic entailment generation, and MRPC for paraphrase generation. These results are crucial for assessing each model's ability to generalize beyond the data types encountered during training. 

In the WikiSmall dataset, Mistral-ReadCtrl outperforms other models with the lowest readability gap of 2.09, suggesting superior control aligning with the target readability levels. It also achieves the highest factuality and coherence scores and significantly outstrips the competition in BLEU and SARI scores, indicating its effectiveness in maintaining content quality in text simplification tasks.

Mistral-ReadCtrl again shows notable performance for the MultiNLI dataset, which focuses on semantic entailment generation, with the lowest readability gap of 2.21 and the highest factuality and coherence scores among the models. However, while its BLEU score is the highest, it slightly trails behind GPT-4 in SARI, demonstrating strong but not leading performance in generating entailments that are semantically aligned with the input.

In the MRPC dataset, which tests the model's ability to generate paraphrases, Mistral-ReadCtrl leads to a readability gap of 1.66, the highest factuality and coherence scores, and the best BLEU and SARI scores. This highlights its exceptional ability to generate paraphrases that not only adhere closely to the desired readability level but also maintain high levels of linguistic and contextual integrity.

Overall, the data from the unseen tasks confirm Mistral-ReadCtrl's robust generalization capabilities across different types of text generation tasks, demonstrating its potential as a versatile tool in NLP applications where adapting to varied content types and maintaining consistent quality is critical.

\begin{figure}
    \vspace{-6mm}
    \centering
    \vspace{-6mm}
    \includegraphics[width=1.0\linewidth]{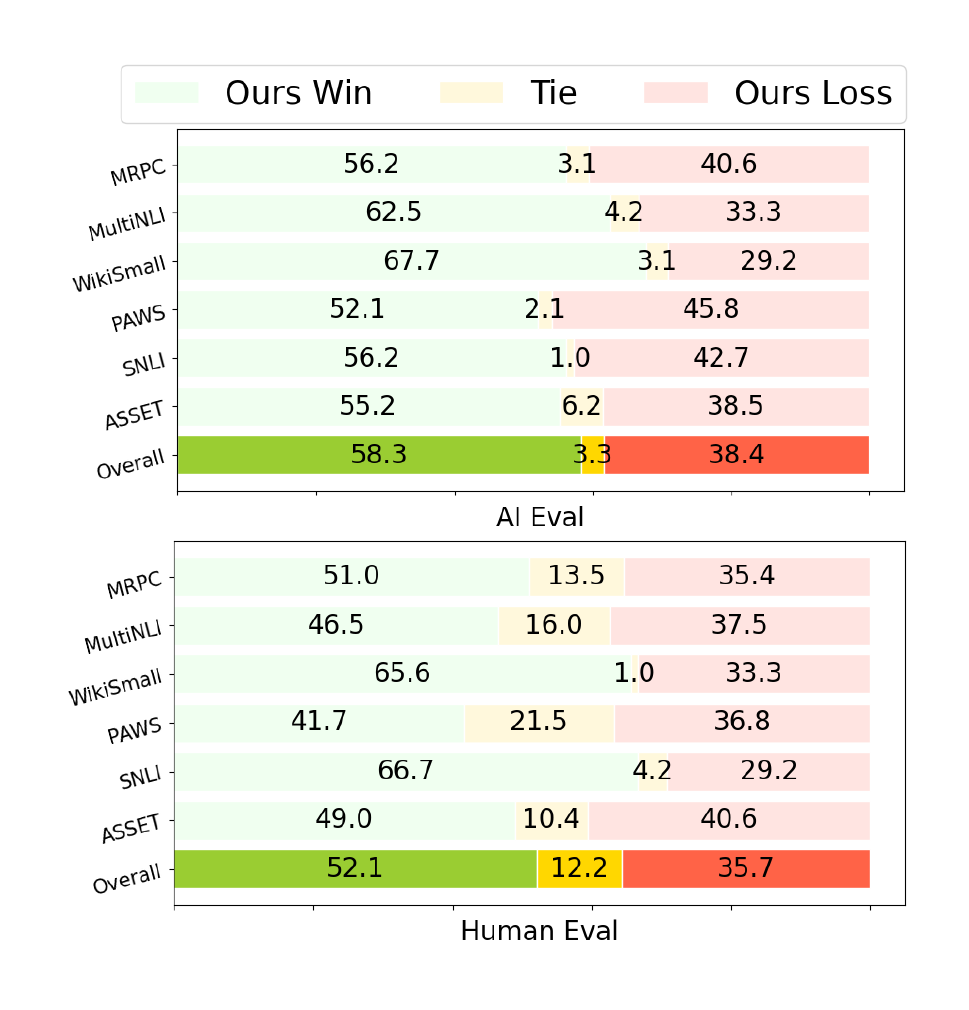}
    \vspace{-12mm}
    \caption{Win rate (\%) for Mistral-ReadCtril vs GPT-4 (3 shots) using AI (Claude3 and GPT3.5) and Human evaluation.}
    \vspace{-6mm}
    \label{fig:human_ai_eval}
\end{figure}

\section{Human Evaluation}
\label{Sec:human_eval}

\subsection{Human Evaluation settings}
Our human evaluation was conducted by 5 human evaluators and 1 expert evaluator~\footnote{More details can be found on Appendix~\ref{apx:human_eval}.}.
We randomly sampled 6 data from the test datasets of 6 data sets, and a total of 36 data appeared in the human evaluation.
We give detailed instructions to the annotators: ``\emph{You are evaluating two systems, both of which are trying to convert inputs to specific readability requirements to produce output suitable for the user. I will show you the input and output of the two systems on grade 2/5/8/11, respectively. Tell me which system's output you prefer by specifying system 1 or system 2 or tie if the quality is the same. Please explain the reason for your preference.}''. And they worked using our evaluation system to select preference; see Figure~\ref{fig:human_eval_system_overview} (left). 
Each time, we randomly shuffle the outputs of two systems (Mistral-ReadCtrl and GPT-4), and they can choose the one that better meets the readability requirements and has higher output quality. If they think the outputs of the two systems are tied, they can choose both.
After we get judgments from multiple people per instance, we do not aggregate their labels before calculating the win rate but count them individually.
We also used a similar setting of our human preference evaluation for AI evaluation with claude-3-opus-20240229 and gpt-3.5-turbo-0125 as the judge~\footnote{More details can be found on Appendix~\ref{apx:ai_eval}}.

After preference evaluation, we then worked with one Linguistics expert for the readability control strategies annotation.
We summarized 4 different reasons for each grade level (see Table~\ref{tab:control_strategy}) and then asked the expert to use our evaluation system for readability control strategies annotation; see Figure~\ref{fig:human_eval_system_overview} (right). 
Each time, the expert needed to select all qualified control strategies for the output of our system (Mistral 7B ReadCtrl), where multiple selections are allowed.

\begin{table}
\vspace{-6mm}
\scriptsize
\begin{tabular}{p{6.2cm}c}
\hline
\textbf{Grade 2}  & 
\\ \hline
Employ short, straightforward sentence structures  & 100\% 
\\ \hline
Focus only on essential details, omitting unnecessary complexity  & 85.7\% 
\\ \hline
Use very simple vocabulary and avoid complex words & 76.2\% 
\\ \hline
Break down information into clear sequential steps & 35.7\% 
\\ \hline

\textbf{Grade 5}  & 
\\ \hline
Introduce some more varied and content-specific vocabulary  & 71.4\% 
\\ \hline
Use longer sentences with conjunctions to combine ideas  & 57.1\% 
\\ \hline
Provide additional context and relevant details & 28.6\% 
\\ \hline
Explain concepts more directly instead of narratives & 23.8\% 
\\ \hline

\textbf{Grade 8}  & 
\\ \hline
Use complex sentence structures like passive voice  & 66.7\% 
\\ \hline
Employ richer descriptive language and vivid details  & 54.8\% 
\\ \hline
Incorporate academic and technical terminology & 47.6\% 
\\ \hline
Establish clear logical connections between ideas & 21.4\% 
\\ \hline

\textbf{Grade 11}  & 
\\ \hline
Construct elaborate compound-complex sentences  & 42.9\% 
\\ \hline
Use sophisticated vocabulary from all domains  & 40.5\% 
\\ \hline
Write with consistent formality and academic tone & 33.3\% 
\\ \hline
Employ advanced stylistic techniques like figurative language & 23.8\%
\\ \hline

\end{tabular}
\vspace{-2mm}
\caption{Readability control strategies for Mistral 7B ReadCtrl. The number represents what proportion of the system output in the corresponding grade level uses the corresponding method to adjust the readability.}
\label{tab:control_strategy}
\vspace{-4mm}
\end{table}

\subsection{Human Evaluation Results}
As shown in Figure~\ref{fig:human_ai_eval}, the human evaluation prefers Mistral-ReadCtrl with an overall win rate of 49.4\% as opposed to GPT-4, while AI evaluation gave us a broader win rate of 58.3\%. 
Specifically, both seen settings (ASSET, SNLI, PAWS) and unseen settings (WikiSmall, MultiNLI, MRPC) exhibit consistent results across human evaluation and AI evaluation. Among these, the lead in WikiSmall and SNLI is most pronounced.
Delving further, human annotations shed light on the operational tactics of Mistral-ReadCtrl. For example, when catering to Grade 2 readability, it implemented straightforward sentence structures with 100\% adherence, focused on essential details 85.7\% of the time, and employed very simple vocabulary in 76.2\% of instances. For more advanced Grade 5 and 8 requirements, it adeptly introduced content-specific vocabulary (71.4\% for Grade 5) and complex sentence structures (66.7\% for Grade 8), illustrating the model's dexterity in scaling complexity according to the readability demands.

\section{Related Work}
Early efforts for readability control in natural language generation (NLG) included microplanning steps to tailor the text to match different target reading levels~\cite{moraes2016enabling,agrawal2019controlling,marchisio2019controlling}. 
More recent studies, such as those by \citet{luo2022readability}, have investigated controllable abstractive and extractive approaches for generating summaries tailored for layman and expert audiences from biomedical documents. 
Concurrently, recent work \citet{pu2023chatgpt,rao2018dear,yao2021improving} examined the ability of the language models to adapt its outputs to different target audiences and writing styles, ranging from formal to informal, whereas \citet{imperial2022uniform} highlighted that GPT2 models struggle with preserving the linguistic complexity of input prompts.
Significant developments have also been made in models for Plain Language Summarization (PLS) from scientific papers~\cite{devaraj2021paragraph,goldsack2023domain,guo2023appls}, focusing on generating simplified texts that retain the original content's meaning. 
Unlike these methods, our novel "readability-controlled instruction learning (ReadCtrl)" method distinctly focuses on fine-grained readability control via direct instruction, allowing for precise adaptation of text complexity. This approach ensures that outputs meet specific readability demands, tested across text simplification, paraphrase generation, and semantic entailment generation tasks. Demonstrating its efficacy and versatility, ReadCtrl performs robustly in both 'seen' and 'unseen' settings.

Text Simplification aims to enhance the readability of sentences by reducing their linguistic complexity, with various important societal applications, such as increasing accessibility for those with cognitive disabilities and also for patient education, non-native speakers, and children with reading difficulties~\cite{martin2020muss,cao2020expertise}.
Various aspects of simplified outputs have been addressed, including 
adapting to specific levels~\cite{nishihara2019controllable},
incorporating edit operations~\cite{kumar2020iterative,mallinson2020felix}, 
enforcing lexical and syntactic constraints~\cite{martin2019controllable}, 
applying linguistically motivated syntactic rules~\cite{maddela2020controllable}, 
and integrating complex span extraction and lay language generation~\cite{chen2018natural, kwon2022medjex,jiang2024medreadme,yao2023readme}
into the simplification process.
In contrast to traditional text simplification, which primarily focuses on controlling the extent of paraphrasing, our approaches are designed to produce succinct and informative output for various tasks in both seen and unseen settings, while maintaining different fine-grained levels of desired readability. Our unique contribution extends readability control beyond mere style transfer to a versatile, instruction-based framework that meets diverse user needs.

Finally, our work follows the instruction tuning technique~\cite{zhang2023instruction} to help LLMs learn to follow readability-controlled instructions. 
Traditional supervised fine-tuning (SFT) techniques often struggle with fine-grained readability control, as they depend on manual annotation or synthetic data generation for enriching readability data, followed by model fine-tuning. 
This method requires considerable financial and time resources, with repeated tasks for each domain-specific application. 
Alternatively, recent advances in instruction learning offer a more generalized approach, as highlighted in several studies~\cite{wei2021finetuned,wang2022self,honovich2022unnatural,zhang2023alpacare,tran2023bioinstruct}.
Instruction learning operates on the premise that the model already possesses the necessary knowledge and skills to perform the target task but requires instructional data to learn how to follow the instructions effectively.
By using a FLAN-style Instruction Fine-Tuning method~\cite{wei2021finetuned}, we gathered task-specific instructions for ReadCtrl and conducted fine-tuning. Our evaluations, both automatic and human, on seen and unseen tasks, confirm ReadCtrl's effectiveness, simplifying the adaptation process and broadening application scope with minimal data needs.

\section{Conclusion}
The ReadCtrl approach significantly enhances the adaptability of LLMs by dynamically adjusting content complexity to suit various readability requirements. Outperforming mainstream models like GPT-4 in evaluations, Mistral-ReadCtrl demonstrates its capability to produce nuanced, high-quality outputs, thereby showing the promise of personalized content generation.

\section{Limitations}
In this paper, we propose a new instruction-learning approach to enhance the controllability of readability levels. While this adjustment is not specific to any particular language, we conducted all of our experiments and analysis exclusively on English-language summarization datasets. 
Additionally, due to the resource limitation, our analysis was limited to Text Simplification (ASSET and WikiSmall datasets), Paraphrase Generation (PAWS and MRPC datasets), and Semantic Entailment Generation (SNLI and MultiNLI datasets), reflecting their prevalent application in NLG studies. Consequently, this paper does not extend to exploring style variations in non-English and other relevant tasks and datasets, such as some mentioned text-to-text generation datasets in the tutorial at ACL 2024~\cite{dou2023automatic}.
Thus, the scope of this study is confined, and the results may not be universally applicable across different linguistic and stylistic contexts.
For readability evaluation, studies have shown that readability formulas may not be ideal for medical text~\cite{zheng2017readability} because short texts (e.g., abbreviations and fragmented texts rather than complete sentences) are common in EHR notes.  In future work, we may explore machine-learning-based approaches~\cite{zheng2018assessing} for readability in subdomains.
Finally, due to resource constraints, we were unable to have actual grade 2, 5, 8, and 11 students provide pairwise preference feedback during our human evaluation. In the future, we plan to collect human evaluation feedback from more appropriate target groups to enhance the reliability of our results further.

\section{Ethics Statement}
While Mistral-ReadCtrl has demonstrated a high level of readability controllability on several NLG datasets dataset, this does not imply their use as general controllable interactive models (like some chatbot systems). These models should be thoroughly evaluated before being used in different settings to ensure reliability.

\bibliography{emnlp24}
\bibliographystyle{acl_natbib}

\newpage

\appendix

\begin{figure*}
    \centering
    \includegraphics[width=1.0\textwidth]{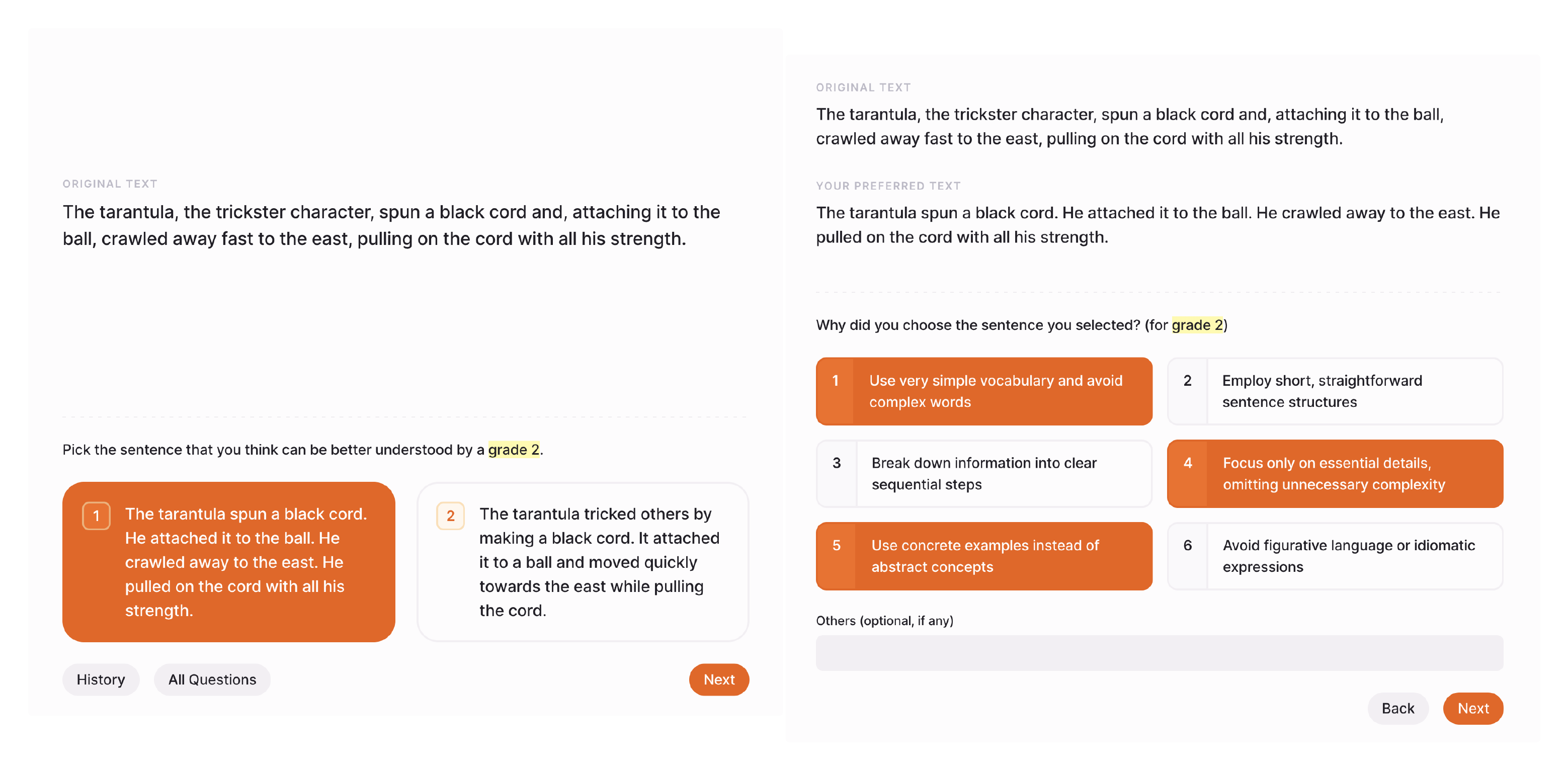}
    \caption{Screenshot of the human evaluation.}
    \label{fig:human_eval_system_overview}
\end{figure*}

\section{Readability Metrics}
\label{apx:read_metrics}

Developed by~\citet{kincaid1975derivation}, the \textbf{Flesch-Kincaid Grade Level(FKGL)} score is a metric that assigns higher scores to texts that are easier to read. It is calculated using the formula:
\[
\tiny \text{FKGL} = 206.835 - 1.015 \left(\frac{\text{totalWords}}{\text{totalSentences}}\right) - 84.6 \left(\frac{\text{totalSyllables}}{\text{totalWords}}\right)
\]

The \textbf{Gunning Fog Index (GFI)}, proposed by~\citet{gunning1952technique}, quantifies the level of formal education required to comprehend a text upon first reading. It is computed as:
\[
\tiny
GFI = 0.4 \left( \frac{\text{totalWords}}{\text{totalSentences}} + 100 \frac{\text{longWords}}{\text{totalWords}} \right)
\]
where \textit{longWords} are defined as words containing more than seven characters. Higher values indicate lower readability.

The \textbf{Automated Readability Index (ARI)}, developed by~\citet{senter1967automated}, correlates to the U.S. school grade level needed to understand the text. It uses the formula:
\[
\tiny
ARI = 4.71 \left(\frac{\text{totalCharacters}}{\text{totalWords}}\right) + 0.5 \left(\frac{\text{totalWords}}{\text{totalSentences}}\right) - 21.43
\]

Developed by~\citet{coleman1975computer}, the \textbf{Coleman-Liau Index (CLI)} focuses on characters rather than syllables to assess text readability. The formula for CLI is:
\[
\tiny
CLI = 0.0588L - 0.296S - 15.8
\]
where \( L \) is the average number of letters per 100 words, and \( S \) is the average number of sentences per 100 words. This metric provides an estimate of the grade level required to understand the text.

\begin{figure}
    \centering
    \vspace{-6mm}
    \includegraphics[width=0.5\textwidth]{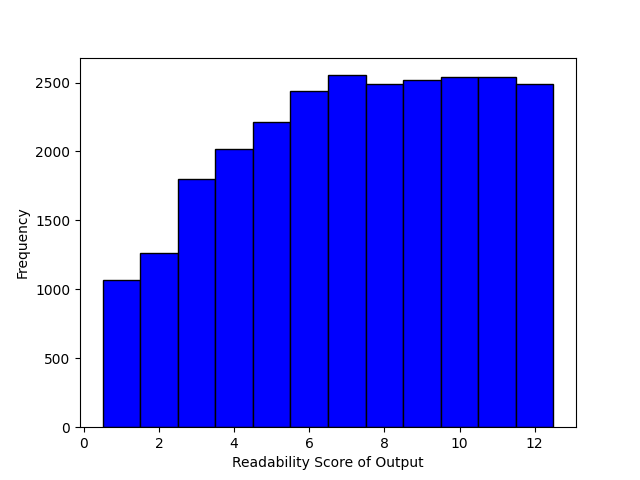}
    \vspace{-6mm}
    \caption{Distribution of examples readability scores from instruction tuning datasets}
    \vspace{-4mm}
    \label{fig:readability_distribution}
\end{figure}

\section{More details about human evaluation}
\label{apx:human_eval}
We provide additional details on our human evaluation setup. Human preference annotators are 5 students who have completed a bachelor's degree or above from an American university and are fluent in English. 
We add some tasks with known answers (i.e., cases where the most/least readable and good/bad quality text should be clear), enabling us to estimate the accuracy of annotators who work on these.  Annotators with low accuracy on tasks with known answers are automatically removed from our worker pool. Only the annotators who passed these final tests were accepted to work on the human preference in this paper.
We gave annotators fair compensation (20\$/hrs).

% We give detailed instructions to the annotators: ``\emph{You are evaluating two systems, both of which are trying to convert inputs to specific readability requirements to produce output suitable for the user. I will show you the input and output of the two systems on grade 2/5/8/11, respectively. Tell me which system's output you prefer by specifying system 1 or system 2 or tie if the quality is the same. Please explain the reason for your preference.}''. And they worked using our evaluation system to select preference; see Figure~\ref{fig:human_eval_system_overview} (left). 
% Each time, we randomly shuffle the outputs of two systems, and they can choose the one that better meets the readability requirements and has higher output quality. If they think the outputs of the two systems are tied, they can choose both.

% We then worked with one Linguistics expert for the readability control strategies annotation.
% We summarized 4 different reasons for each grade level (see Table~\ref{tab:control_strategy}) and then asked the expert to use our evaluation system for readability control strategies annotation; see Figure~\ref{fig:human_eval_system_overview} (right). 
% Each time, the expert needed to select all qualified control strategies for the output of our system (Mistral 7B ReadCtrl), where multiple selections are allowed.

\section{More details about LLM evaluation}
\label{apx:ai_eval}

To reduce the heavy human evaluation and make the evaluation easier to reproduce, we use a similar setting of our human preference evaluation for AI evaluation. 
Comparison-based feedback evaluation assesses the accuracy of LLM in deciding preferences between two responses. 
However, it is widely acknowledged that current LLMs exhibit significant \textbf{positional bias} \cite{lan2024criticbench,wang2023large,zheng2024judging,zeng2023evaluating}, i.e., LLMs tend to prefer responses based on their specific position in the prompt. 
We implement a rigorous verification process to mitigate the effects of positional bias to evaluate the real capability. Specifically, given responses \(R_a\) and \(R_b\) to be compared, we obtain the comparison based on two orders, noted as \(F^c_a = F_c(R_a, R_b)\) and \(F^c_b = F_c(R_b, R_a)\). The objective scores are computed by:
\[
s = \frac{1}{N} \sum_{i=1}^N 1(L(F^c_{a,i}, F^c_{b,i}))
\]
where \(L(F^c_a, F^c_b)\) is true if and only if \(F^c_a \neq F^c_b\) and \(F^c_a, F^c_b\) align with ground-truth preference label. \(N\) is the number of test samples.
The prompts we used for LLM-as-a-judge (claude-3-opus-20240229 and gpt-3.5-turbo-0125) evaluation can be found in Table 4.

\begin{table}
\scalebox{0.8}{
    \centering
    \begin{tabular}{lc}
        \toprule
        \textbf{Parameter}             & \textbf{Value} \\
        \midrule
        Computing Infrastructure       & 40GB NVIDIA A100 GPU \\
        Optimizer                      & Adam \\
        Optimizer Params               & $\beta = (0.9, 0.999), \epsilon = 10^{-8}$ \\
        Learning rate                  & $3 \times 10^{-4}$ \\
        Learning Rate Decay            & Linear \\
        Weight Decay                   & 0 \\
        Warmup Steps                   & 200 \\
        Batch size                     & 128 \\
        Epoch                          & 5 \\
        \bottomrule
    \end{tabular}
}
    \caption{Hyperparameter settings for Mistral 7B ReadCtrl.}
    \label{tab:hyperparameters}
\end{table}

\section{Hyper-parameter Settings}
The experiments were executed using the version 4.37.1 of the transformers library released by Hugging Face. In Table 3, we report the hyperparameters used to train the models on our combined dataset. We use the Adam optimizer and employ a linearly decreasing learning rate schedule with warm-up step is 200. In this section, we detail our experimental setup, the datasets employed, and the evaluation strategy adopted for assessing the performance of our instruction-tuned LLMs in various BioNLP tasks. Furthermore, all experiments were conducted using two Nvidia A100 GPUs, each with 40 GB of memory. The CPU used was an Intel Xeon Gold 6230 processor, and the system was equipped with 192 GB of RAM.

\section{Experiments with GPT3.5, GPT4, Claude-3}
All of our experiments were conducted on the version of GPT3.5, GPT4 and Claude 3 between 25 March 2023 and 13 April 2024 by using the OpenAI’s API.10 We set temperature = 1, top\_p=1, frequency penalty = 0, and presence penalty = 0.

\section{ReadCtrl instruction following evaluation setting}
\label{apx:ReadCtrl_following_setting}
We have plotted Figures 1, 6, 7, 8, 9, 10, and 11 by calculating the readability scores or reading levels of the outputs generated in response to prompts that request specific reading levels ranging from 1 to 12. These calculations were performed across all test sets of the six datasets mentioned in the Experiment section. Additionally, we calculated the standard deviation of the readability scores across these test sets to assess the consistency of the output's readability.

\section{Examples of output generated by Mistral-ReadCtrl and GPT4 during ReadCtrl instruction following evaluation}
\label{apx:ReadCtrl_following_setting}

Tables~\ref{tab:example} present distinct levels of output generated by the Mistral-ReadCtrl and GPT4 and their readability scores given by Flesch-Kincaid Grade Level (FKG), Gunning fog index (GFI), and Coleman-Liau index (CLI) metrics.

We will delve into the observed discrepancies between the Readability Gap and the performance curves in our evaluation, as demonstrated by our results for the PAWS and MultiNLI datasets. The Readability Gap, calculated as the average difference between the actual readability score of the output and the requested readability score across all samples, shows intriguing variations in behavior across different datasets.

For the PAWS dataset, although the Readability Gap appears almost perfect in Table 1, the corresponding curve does not exhibit as favorable performance. This anomaly can be attributed to the output readability distribution of PAWS, which is somewhat concentrated within a specific range (typically between 4-8). While this concentration allows for excellent performance within this median range, it leads to a less generalized performance across the full spectrum of readability levels (from 1-12). Therefore, even a small Readability Gap in numerical terms may not accurately reflect an evenly distributed ability to target all requested readability levels.

Conversely, the MultiNLI dataset exhibits a larger Readability Gap in Table 1, yet the performance curve approaches perfection. This suggests that while the average gap is larger, the outputs are more uniformly distributed across the entire range of readability levels, allowing for closer adherence to the target levels across a broader spectrum. This indicates a more generalized and adaptable performance despite the numerically larger gap.

This analysis underscores the importance of considering both the Readability Gap and the distribution of output readability scores when assessing model performance. A low Readability Gap might suggest excellent average performance but could conceal poor adaptability across a range of readability levels. Conversely, a higher Readability Gap might indicate a more uniform distribution of performance across all levels, suggesting a different kind of effectiveness.

Further investigation into these patterns for all six datasets employed in our study reveals similar trends.

\begin{figure}
    \centering
    \includegraphics[width=1.0\linewidth]{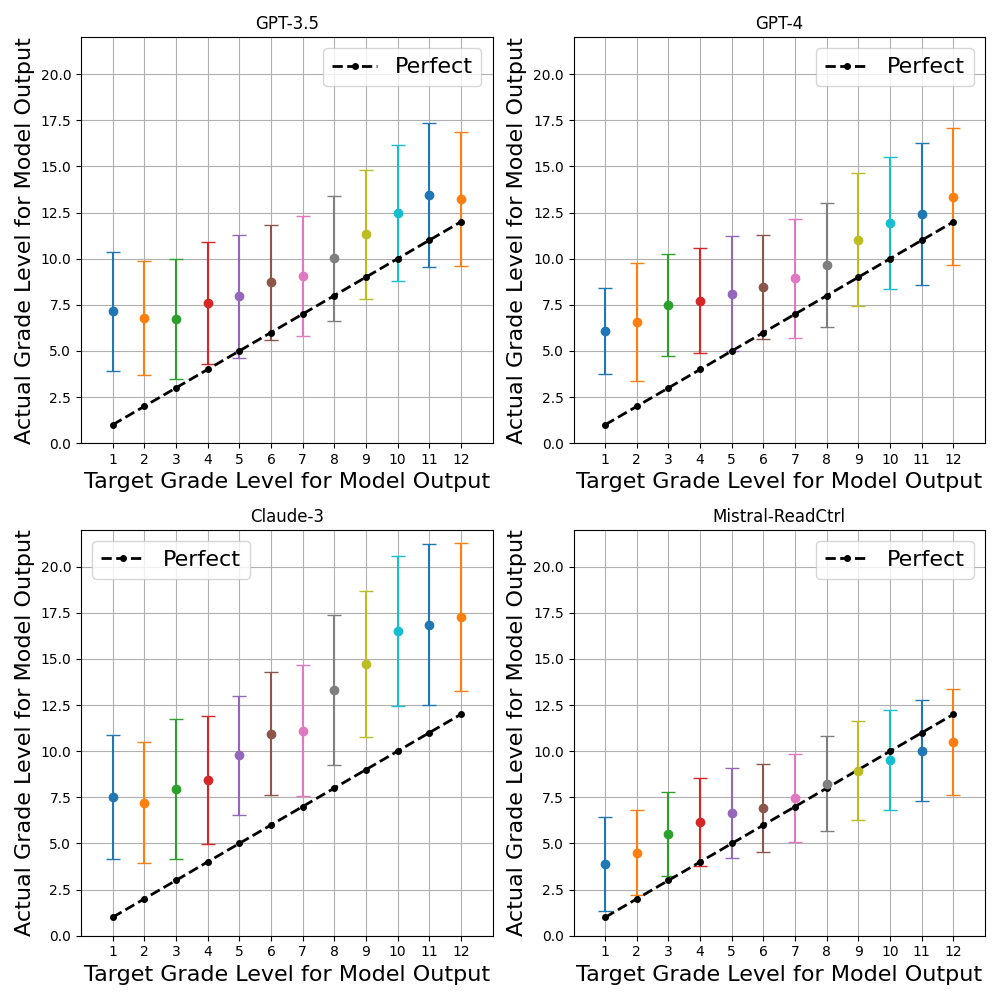}
    \caption{ReadCtrl instruction following ability evaluation on ASSET.}
    \label{fig:asset_full}
\end{figure}

\begin{figure}
    \centering
    \includegraphics[width=1.0\linewidth]{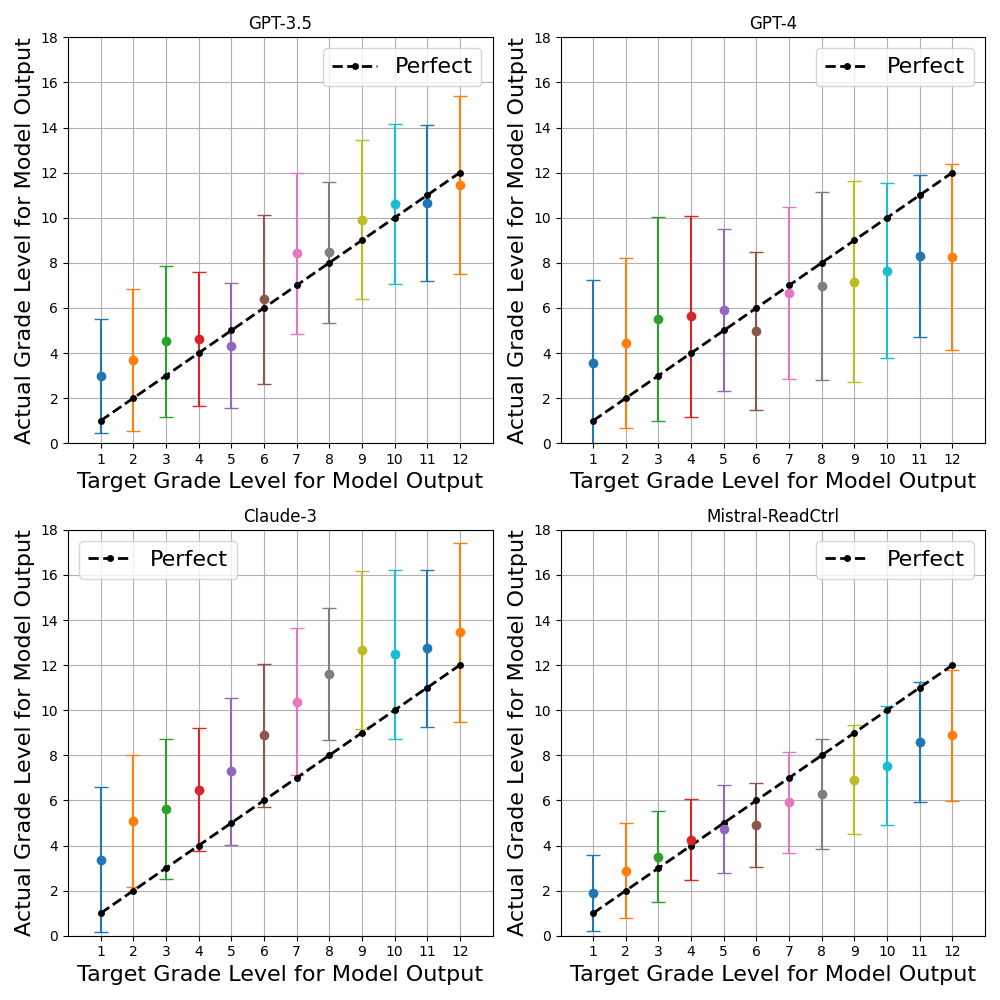}
    \caption{ReadCtrl instruction following ability evaluation on SNLI}
    \label{fig:snli_full}
\end{figure}

\begin{figure}
    \centering
    \includegraphics[width=1.0\linewidth]{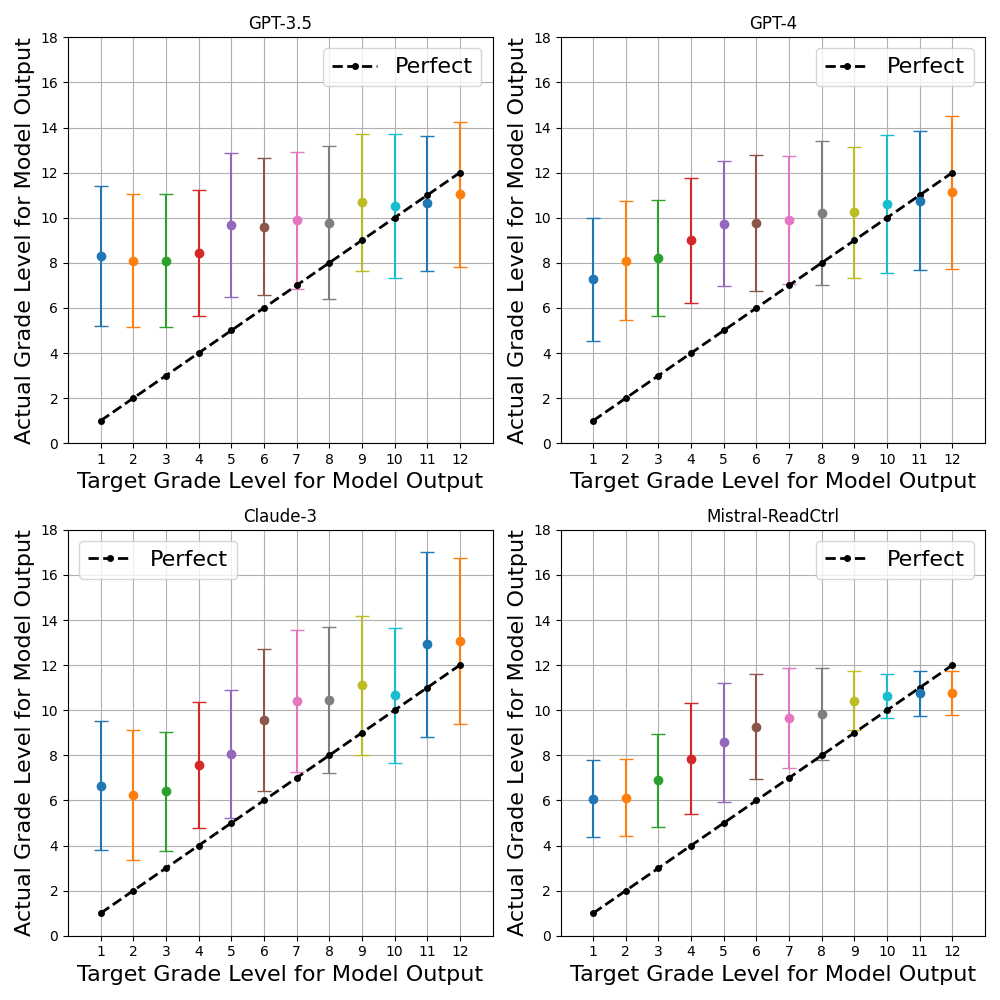}
    \caption{ReadCtrl instruction following ability evaluation on PAWS}
    \label{fig:paws_full}
\end{figure}

\begin{figure}
    \centering
    \includegraphics[width=1.0\linewidth]{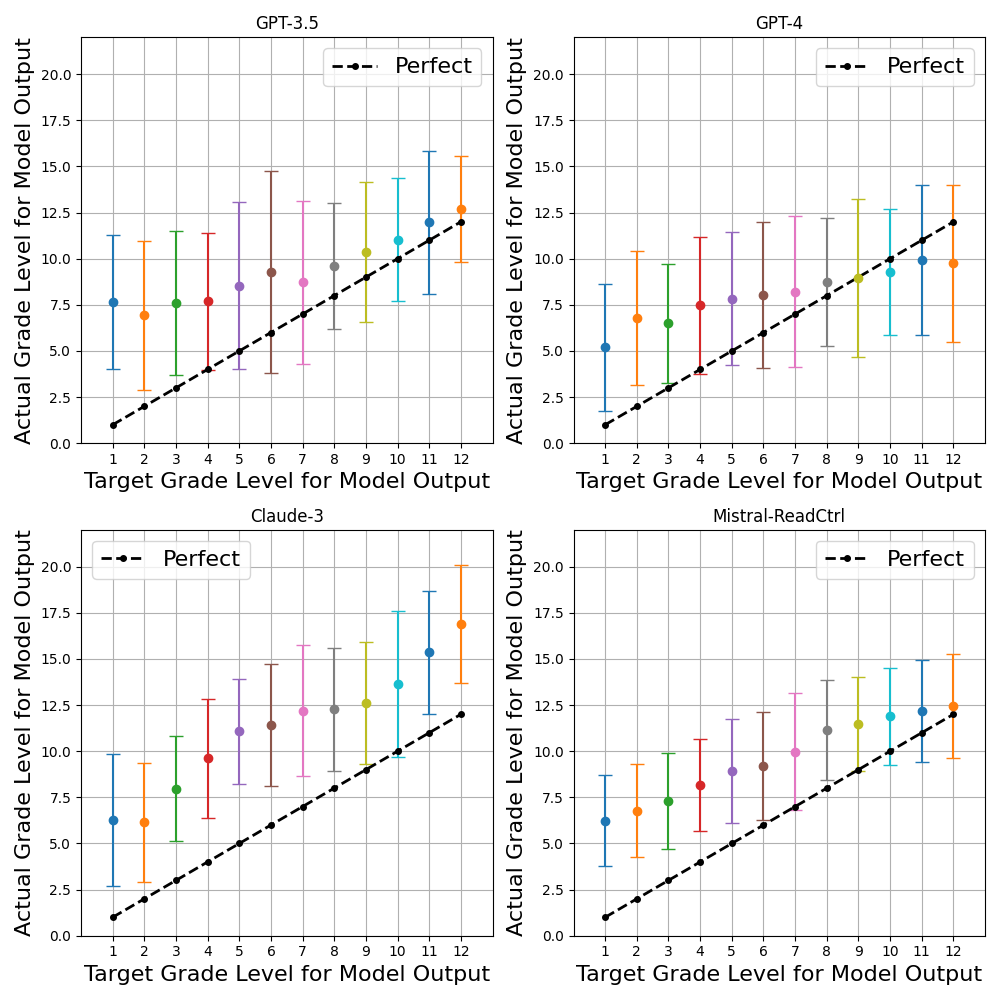}
    \caption{ReadCtrl instruction following ability evaluation on WikiSmall}
    \label{fig:wikismall_full}
\end{figure}

\begin{figure}
    \centering
    \includegraphics[width=1.0\linewidth]{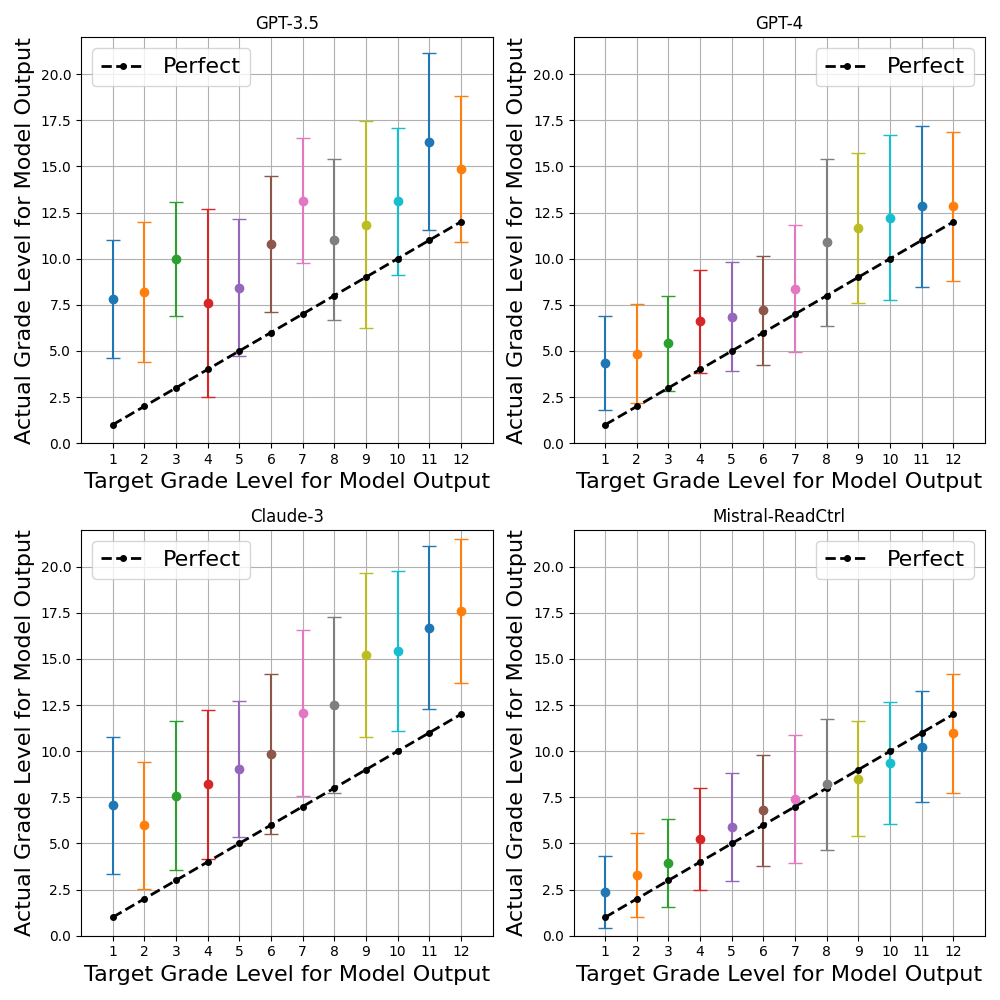}
    \caption{ReadCtrl instruction following ability evaluation on MultiNLI}
    \label{fig:multinli_full}
\end{figure}

\begin{figure}
    \centering
    \includegraphics[width=1.0\linewidth]{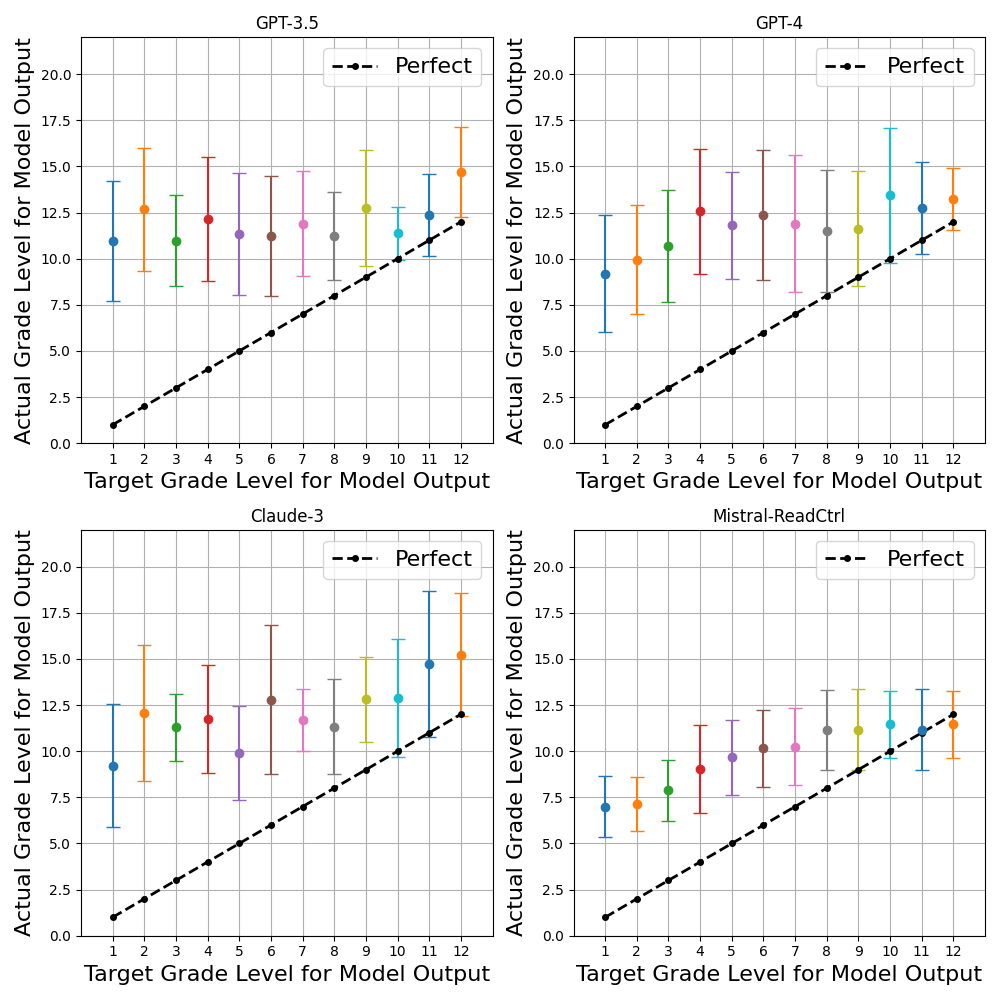}
    \caption{ReadCtrl instruction following ability evaluation on MRPC}
    \label{fig:mrpc_full}
\end{figure}

\onecolumn
\section{Prompts}
\label{prompts}

\lstset{
    basicstyle=\ttfamily,
    columns=fullflexible,
    breaklines=true,
    postbreak=\mbox{\textcolor{red}{$\hookrightarrow$}\space}
}
{\tiny
% \begin{table}%[ht]
% \caption{}
% \label{prompts}
\centering

\begin{tabular}{|p{1cm}|p{13.5cm}|}
\hline
\textbf{Type}  & \textbf{Prompt} \\ \hline
AI Evaluation & 
\begin{lstlisting}
You are evaluating two systems, both of which are trying to convert inputs to specific readability requirements to produce output suitable for the user.
I will show you the input and output of the two systems on grade 2/5/8/11, respectively. Tell me which system's output you prefer by specify system 1 or system 2 or tie if the quality is the same. Please explain the reason for your preference.
Input:
{input}
System 1 output:
grade 2:
{system1_2}
grade 5:
{system1_5}
grade 8:
{system1_8}
grade 11:
{system1_11}
System 2 output:
grade2:
{system2_2}
grade 5:
{system2_5}
grade 8:
{system2_8}
grade 11:
{system2_11}

Please use the following json format for your output:
'grade 2 preference': xxxx,
'grade 2 preference reasons': xxxx,
'grade 5 preference': xxxx,
'grade 5 preference reasons': xxxx,
'grade 8 preference': xxxx,
'grade 8 preference reasons': xxxx,
'grade 11 preference': xxxx,
'grade 11 preference reasons': xxxx
Please only output your response following the required format, and do not output any other content. Now tell me your preference and reasons:

\end{lstlisting}
\\ \hline
Readability Control Generation Prompt & 
\begin{lstlisting}
Below is an instruction that describes a task. Write a response that appropriately completes the request.
### Instruction:
please output an entailment at a {grade}-reading level.
### Input:
{input}
### Response:
\end{lstlisting}
\\ \hline

\end{tabular}

\small
Table 4: All prompts used in our work.

\begin{table*}
    \centering
    \begin{tabular}{>{\raggedright\arraybackslash}p{11cm} c c c}
        \toprule
        \textbf{Document} & \textbf{FKG} & \textbf{GFI}& \textbf{CLI} \\
        \midrule
The tarantula, the trickster character, spun a black cord and, attaching it to the ball, crawled away fast to the east, pulling on the cord with all his strength. & 9.9 & 15.74 & 14.8 \\
        \midrule
        \textbf{Requested level: second-grade reading level(Mistral ReadCtrl)} \\
        The tarantula spun a black cord. He attached it to the ball. He crawled away to the east. He pulled on the cord with all his strength. & 0.0 & 4.2 & 0.8 \\
        \textbf{Requested level: second-grade reading level(GPT-4)} \\
        The tarantula tricked others by making a black cord. It attached it to a ball and moved quickly towards the east while pulling the cord. & 4.6 & 6.6 & 5.9 \\

        \midrule
        
        \textbf{Requested level: fifth-grade reading level(Mistral ReadCtrl)} \\
        The tarantula, the trickster character, spun a black cord. He attached it to the ball and crawled away fast to the east. He pulled on the cord with all his strength. & 2.6 & 6.7 & 4.2 \\
        \textbf{Requested level: fifth-grade reading level(GPT-4)} \\
        The tarantula, who is known for playing tricks, made a black cord. He attached it to the ball and quickly crawled east, pulling the cord as hard as he could. & 4.4 & 7.33 & 6.2 \\

        \midrule
        
        \textbf{Requested level: eighth-grade reading level(Mistral ReadCtrl)} \\
        The tarantula, the trickster character, spun a black cord and attached it to the ball. He crawled away fast to the east, pulling on the cord with all his strength. & 4.4 & 8.67 & 7.1 \\

         \textbf{Requested level: eighth-grade reading level(GPT-4)} \\
         The tarantula, the sneaky character, made a black cord and tied it to the ball. It quickly moved east, pulling the cord strongly. & 4.2 & 8.08 & 6.2 \\

        \midrule
        
        \textbf{Requested level: eleventh-grade reading level(Mistral ReadCtrl)}\\
        The tarantula, the trickster character, spun a black cord and attached it to the ball before crawling away to the east, pulling on the cord with all his strength. & 11.1 & 14.36 & 14.8 \\

        \textbf{Requested level: eleventh-grade reading level(GPT-4)} \\ 
        The character of the tarantula, known for trickery, spun a black cord. He attached this cord to the ball and rapidly scuttled eastward, tugging at the cord with maximum force. & 6.8 & 10.0 & 9.0 \\
        
        \bottomrule
    \end{tabular}
    \caption{Examples of generated summaries for different readability levels measured using FKG, GFI and CLI metrics.}
    \label{tab:example}
\end{table*}

\end{document}